\pdfoutput=1

\documentclass[11pt]{article}
\usepackage[utf8]{inputenc}

\usepackage[final]{acl}

\usepackage[utf8]{inputenc}
\usepackage[T1]{fontenc}
\usepackage{hyphenat}
\usepackage{xspace}
\usepackage{amsmath}
\usepackage{amsfonts}
\usepackage{hyperref}
\usepackage{url}
\usepackage{booktabs}
\usepackage{multirow}
\usepackage{makecell}
\usepackage{caption}
\usepackage{minibox}
\usepackage{bbm}
\usepackage{graphicx}
\usepackage{balance}
\usepackage{mathtools}
\usepackage{color}
\usepackage{marvosym}
\usepackage{ifthen}
\usepackage{textcomp}
\usepackage{enumitem}
\usepackage{verbatim}
\usepackage{algorithm}
\usepackage{numprint}
\usepackage{balance}

\usepackage{amsthm}
\theoremstyle{plain}

\newcommand{\chatoDisplayMode}[1]{#1}

\definecolor{MyRed}{rgb}{0.6,0.0,0.0} 
\definecolor{MyBlack}{rgb}{0.1,0.1,0.1} 
\newcommand{\inred}[1]{{\color{MyRed}\sf\textbf{\textsc{#1}}}}
\newcommand{\frameit}[2]{
  \begin{center}
  {\color{MyRed}
  \framebox[.9\columnwidth][l]{
    \begin{minipage}{.85\columnwidth}
    \inred{#1}: {\sf\color{MyBlack}#2}
    \end{minipage}
  }\\
  }
  \end{center}
}

\newcommand{\note}[2][]{\chatoDisplayMode{\def\@tmpsig{#1}\frameit{{\Pointinghand} Note}{#2\ifx \@tmpsig \@empty \else \mbox{ --\em #1}\fi}}}
\newcommand{\todo}[2][]{\chatoDisplayMode{\def\@tmpsig{#1}\frameit{{\Writinghand} To-do}{#2\ifx \@tmpsig \@empty \else \mbox{ --\em #1}\fi}}}

\newcommand{\abbrevStyle}[1]{#1}

\newcommand{\ie}{\abbrevStyle{i.e.}\xspace}
\newcommand{\eg}{\abbrevStyle{e.g.}\xspace}
\newcommand{\cf}{\abbrevStyle{cf.}\xspace}

\newcommand{\vs}{\abbrevStyle{vs.}\xspace}

\newcommand{\Secref}[1]{Sec.~\ref{#1}}

\newcommand{\Tabref}[1]{Table~\ref{#1}}
\newcommand{\Figref}[1]{Fig.~\ref{#1}}

\newcommand{\Appref}[1]{Appendix~\ref{#1}}

\newcommand{\xhdr}[1]{\vspace{1.7mm}\noindent{{\bf #1.}}}

\newcommand{\textcite}[1]{\citeauthor{#1} \shortcite{#1}}

\newcommand{\hide}[1]{}

\makeatletter
\newcommand{\iffont}[2]{\ifthenelse{\equal{\f@family}{#1}}{#2}{}}
\makeatother

  \usepackage{mathptmx}

  \DeclareSymbolFont{greek}{OML}{cmm}{m}{n}
  \DeclareMathSymbol{\alpha}{\mathalpha}{greek}{"0B}
  \DeclareMathSymbol{\beta}{\mathalpha}{greek}{"0C}
  \DeclareMathSymbol{\gamma}{\mathalpha}{greek}{"0D}
  \DeclareMathSymbol{\delta}{\mathalpha}{greek}{"0E}
  \DeclareMathSymbol{\epsilon}{\mathalpha}{greek}{"0F}
  \DeclareMathSymbol{\zeta}{\mathalpha}{greek}{"10}
  \DeclareMathSymbol{\eta}{\mathalpha}{greek}{"11}
  \DeclareMathSymbol{\theta}{\mathalpha}{greek}{"12}
  \DeclareMathSymbol{\iota}{\mathalpha}{greek}{"13}
  \DeclareMathSymbol{\kappa}{\mathalpha}{greek}{"14}
  \DeclareMathSymbol{\lambda}{\mathalpha}{greek}{"15}
  \DeclareMathSymbol{\mu}{\mathalpha}{greek}{"16}
  \DeclareMathSymbol{\nu}{\mathalpha}{greek}{"17}
  \DeclareMathSymbol{\xi}{\mathalpha}{greek}{"18}
  \DeclareMathSymbol{\pi}{\mathalpha}{greek}{"19}
  \DeclareMathSymbol{\rho}{\mathalpha}{greek}{"1A}
  \DeclareMathSymbol{\sigma}{\mathalpha}{greek}{"1B}
  \DeclareMathSymbol{\tau}{\mathalpha}{greek}{"1C}
  \DeclareMathSymbol{\upsilon}{\mathalpha}{greek}{"1D}
  \DeclareMathSymbol{\phi}{\mathalpha}{greek}{"1E}
  \DeclareMathSymbol{\chi}{\mathalpha}{greek}{"1F}
  \DeclareMathSymbol{\psi}{\mathalpha}{greek}{"20}
  \DeclareMathSymbol{\omega}{\mathalpha}{greek}{"21}
  \DeclareMathSymbol{\varepsilon}{\mathalpha}{greek}{"22}
  \DeclareMathSymbol{\vartheta}{\mathalpha}{greek}{"23}
  \DeclareMathSymbol{\varpi}{\mathalpha}{greek}{"24}
  \DeclareMathSymbol{\varrho}{\mathalpha}{greek}{"25}
  \DeclareMathSymbol{\varsigma}{\mathalpha}{greek}{"26}
  \DeclareMathSymbol{\varphi}{\mathalpha}{greek}{"27}
  \DeclareSymbolFont{otone}{OT1}{cmr}{m}{n}
  \DeclareMathSymbol{\Gamma}{\mathalpha}{otone}{0}
  \DeclareMathSymbol{\Delta}{\mathalpha}{otone}{1}
  \DeclareMathSymbol{\Theta}{\mathalpha}{otone}{2}
  \DeclareMathSymbol{\Lambda}{\mathalpha}{otone}{3}
  \DeclareMathSymbol{\Xi}{\mathalpha}{otone}{4}
  \DeclareMathSymbol{\Pi}{\mathalpha}{otone}{5}
  \DeclareMathSymbol{\Sigma}{\mathalpha}{otone}{6}
  \DeclareMathSymbol{\Upsilon}{\mathalpha}{otone}{7}
  \DeclareMathSymbol{\Phi}{\mathalpha}{otone}{8}
  \DeclareMathSymbol{\Psi}{\mathalpha}{otone}{9}
  \DeclareMathSymbol{\Omega}{\mathalpha}{otone}{10}
  \DeclareSymbolFont{syms}{OML}{cmm}{m}{it}
  \DeclareMathSymbol{\partial}{\mathord}{syms}{"40}
  \DeclareMathAlphabet{\mathbold}{OML}{cmm}{b}{it}
  \DeclareSymbolFont{largesymbols}{OMX}{cmex}{m}{n}

  \DeclareMathAlphabet{\mathcal}{OMS}{cmsy}{m}{n}

\usepackage{times}
\usepackage{latexsym}

\usepackage[T1]{fontenc}

\usepackage[utf8]{inputenc}

\usepackage{microtype}

\usepackage{inconsolata}

\usepackage{graphicx}

\title{Combining Constrained and Unconstrained Decoding via Boosting:\\
BoostCD and Its Application to Information Extraction}

\author{Marija \v{S}akota  \quad Robert West\\
EPFL, Lausanne, Switzerland \\
\texttt{\{marija.sakota, robert.west\}@epfl.ch}\\
}

\newcommand{\model}{BoostIE}
\newcommand{\method}{BoostCD}
\newcommand{\relik}{ReLiK}
\newcommand{\synthiedata}{Wiki-cIE Code}

\begin{document}
\maketitle

\begin{abstract}
Many recent approaches to structured NLP tasks use an autoregressive language model $M$ to map unstructured input text $x$ to output text $y$ representing structured objects (such as tuples, lists, trees, code, etc.), where the desired output structure is enforced via constrained decoding.
During training, these approaches do not require the model to be aware of the constraints, which are merely implicit in the training outputs $y$.
This is advantageous as it allows for dynamic constraints without requiring retraining, but can lead to low-quality output during constrained decoding at test time.
We overcome this problem with \textit{Boosted Constrained Decoding (\method{}),} which combines constrained and unconstrained decoding in two phases:
Phase~1 decodes from the base model $M$ twice, in constrained and unconstrained mode, obtaining two weak predictions.
In phase~2, a learned autoregressive boosted model combines the two weak predictions into one final prediction.
The mistakes made by the base model with \vs\ without constraints tend to be complementary, which the boosted model learns to exploit for improved performance.
We demonstrate the power of \method{} by applying it to closed information extraction.
Our model, \textit{\model{},} outperforms prior approaches both in and out of distribution, addressing several common errors identified in those approaches.

\end{abstract}

\section{Introduction}
\begin{figure}[t]
    \centering
    \includegraphics[width=
    \columnwidth]{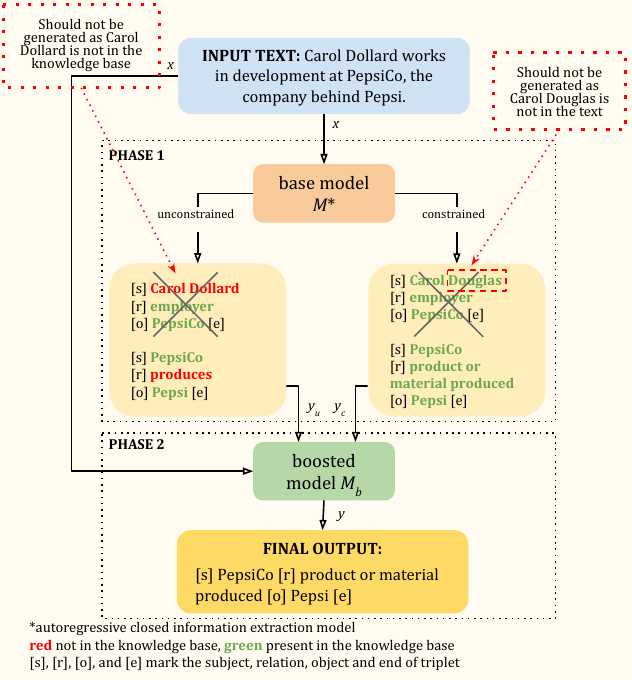}
    \caption{
    \textbf{Overview of \method{},}
    exemplified on the task of closed information extraction (\model{}).
    Phase~1 applies the base model twice on input $x$: unconstrained and constrained.
    Phase~2 combines the two resulting weak predictions $y_u$ and $y_c$ into final prediction $y$ using a boosted model, which during training learns to undo mistakes made by the base model.
    }
    \label{fig:diagram}
\end{figure}

Extracting structured semantic information from unstructured text is essential for many AI tasks, including knowledge discovery \cite{ji-grishman-2011-knowledge}, knowledge base maintenance \cite{tang-etal-2019-learning}, symbolic representation, reasoning \cite{9416312}, and planning. Beyond these applications, a growing number of NLP tasks now explicitly require structured outputs as part of their formulation. Some examples are code generation \cite{poesia2022synchromeshreliablecodegeneration}, SQL generation \cite{scholak-etal-2021-picard}, constituency parsing \cite{deutsch-etal-2019-general}, and various information extraction \cite{decao2021autoregressiveentityretrieval, synthie, relik}.

Many recent approaches for these tasks use autoregressive models trained on pairs of unstructured input text and structured output targets, coupled with constrained decoding \cite{synthie, genie, webie, decao2021autoregressiveentityretrieval}. In real-world tasks, the constraints can often change, so constrained decoding offers an easy way to adapt the schema without the need to retrain the model. Constrained decoding also helps steer the model the right way when it is already close to generating the correct output (\eg, when the only problems are minor surface form discrepancies). However, on the downside, as the model remains unaware of the explicit constraints until decoding at inference time, it may generate less plausible outputs when the input data or the constraints at inference time deviate from those seen during training.

We illustrate in \Figref{fig:diagram}, which shows an example of outputs of the autoregressive model with constrained decoding on the closed information extraction (cIE) task, where the goal is to extract complete sets of fact triplets (subject, relation, object) from text, where all entities and relations must be present in a predefined knowledge base (KB). In the provided example, the base model is a cIE model trained on exhaustive data (\ie, facts in the text are fully expressible under KB constraints). The shown input text differs from the training data by containing facts that are not expressible under KB constraints. The base model generates two triplets when run in unconstrained mode. For the first one, the entity present in the text, ``Carol Dollard'', is not present in the KB. Because the base model was trained on exhaustive data, when prompted in an unconstrained manner, it generates a correct triplet that captures this entity. When constrained, however, instead of removing this triplet entirely (as it does not comply with the KB), the model resorts to generating a triplet with a wrong entity with a similar name (``Carol Douglas'').
For the second triplet, the unconstrained model generates correct entities but makes a formatting error in the relation (``produces'' instead of ``product or material produced''). In this case, constrained decoding helps by correcting the relation name.
Ideally, we seek a method able to recognize patterns in the constrained and unconstrained outputs to combine their strengths and recover from their errors, without having to know the explicit constraints already at training time (which would reduce flexibility as constraints change, \eg, as the KB evolves).

To overcome these problems, we introduce \textit{Boosted Constrained Decoding (\method{}),} a method
with the ability to correct
systematic errors that an autoregressive model trained for a structured NLP task might make during constrained as well as unconstrained generation.
\method{} works in two phases:
Phase~1 decodes from the base model $M$ twice for the input text $x$, in constrained and unconstrained mode, obtaining two weak predictions $y_c$ and $y_u$.
In phase~2, a learned autoregressive boosted model combines the two weak predictions into one final prediction $y$.
Empirically, the mistakes made by the base model with \vs\ without constraints tend to be complementary, which the boosted model learns to exploit during training for improved performance.

To demonstrate the power of the \method{} paradigm, we apply it to closed information extraction (cIE; \cf\ \Figref{fig:diagram}) as an example of a structured task with constraints (defined by the content of the knowledge base) that tend to change dynamically in real-life settings.
We further enhance the resulting cIE model, \textit{\model{},} with Direct Preference Optimization (DPO) \cite{rafailov2023direct} for improving performance on out-of-distribution data.
We show that \model{} outperforms previous methods both in-distribution (on synthetic data it was trained on) and out-of-distribution (on random Wikipedia paragraphs). We also demonstrate that \model{} lowers the rate of common errors made by earlier techniques.

\xhdr{Contributions}
Our contributions are as follows:


(i) We propose \method{}, a new method for training autoregressive language models for structured NLP tasks.

(ii) We instantiate \method{} for the closed information extraction (cIE) task, obtaining a model termed \textit{\model{},} and conduct a detailed evaluation, showing that \model{} outperforms existing methods both in-distribution (by 17.05 and 12.56 absolute points in micro and macro F1, respectively) and out-of-distribution (by 10.94 and 12.54 absolute points in micro and macro F1, respectively).

(iii) A detailed error analysis confirms that \model{} lowers the rate of common errors made by previous cIE models, as well as disadvantages of vanilla constrained decoding for this task.

(iv) We share our code, models, and data for researchers to reuse and extend: \url{https://github.com/epfl-dlab/BoostCD}

\section{Method: \method}
\label{sec:method}

Language models trained for structured NLP tasks in a supervised manner can often perform reasonably well even without the constraints imposed,
but the constraints are still required to guarantee 100\% valid generations, and they can steer the model to pick the one correct output when multiple outputs might seem plausible \textit{a priori} (\eg, when an entity has multiple aliases).
However, when the constraints require altering the unconstrained output significantly (\eg, when an entity generated in unconstrained mode is not present in the KB), performance can suffer from imposing the constraints (for a more formal evaluation of this, see analysis in \Appref{appendix:constrained_unconstrained_analysis}).


We hence seek a method that enjoys the benefits of constraints without suffering from their negative side effects.
In developing such a method, we draw inspiration from boosting \cite{boosting}, a classic ensemble learning technique that aims to improve performance by iteratively combining weaker models into a single stronger one. The idea is to train models sequentially, where each new model focuses on the mistakes made by the previous ones. The final prediction is formed by aggregating the outputs of all models, often through weighted voting or summation.
Our method, \textit{Boosted Constrained Decoding (\method{}),} trains a new model, the \textit{boosted model,} to predict the ground-truth output based on both the constrained and the unconstrained generation from the autoregressive base model together with the input text.
This way of training allows the boosted model to recover from 
systematic mistakes made by the base model without requiring explicit knowledge of the constraints at training time.

For intuition, consider the cIE task as illustrated in \Figref{fig:diagram}:
in the example, unconstrained decoding extracted a triplet (Carol Dollard, employer, PepsiCo) that was not extracted by constrained decoding (because Carol Dollard is not in the KB);
and constrained decoding extracted a triplet (Carol Douglas, employer, PepsiCo) that was not extracted by unconstrained decoding (because Carol Douglas is not mentioned in the input text).
By seeing such candidate triplets together with the ground-truth triplet set (which contains neither of the above candidate triplets), the boosted model learns to recognize that entities occurring only in the constrained but not the unconstrained output (or \textit{vice versa}) indicate triplets that were erroneously extracted by the base model and should thus be discarded.
Note that this is but one of the many potential patterns that the boosted model might learn.

\xhdr{Pipeline}
The \method\ pipeline is shown in \Figref{fig:diagram}. For illustration, we use the example of cIE, although our method can be applied to any structured extraction task. Under the assumption that we have a dataset which consists of pairs $(x,y)$, where $x$ is the input text, and $y = \{(s,r,o)|(s,r,o)\in E\times R \times E \}$ (a set of triplets constrained to the KB that contains all entities $E$ and relations $R$), our training pipeline consists of two phases:

(i) \textbf{Phase 1:} We use a base model $M$, trained in an autoregressive manner on $(x,y)$ pairs, to make two parallel passes. In one pass, we let the model generate in an unconstrained manner: by sending input text $x$ to the model $M$ without imposing any constraints, we obtain the output $y_u$. In the other pass, we generate by imposing constraints: by providing the input text $x$ and using $M$ with constrained decoding, we obtain the output $y_c$.

(ii) \textbf{Phase 2:} In this phase, we train the boosted model $M_b$ to correct the errors that the base model $M$ made in phase~1. $M_b$ is trained in an autoregressive way
to map $(x, y_u, y_c)$ (\ie, the original input together with both phase-1 predictions) to the ground-truth output $y$.

During the inference, we repeat the steps from both phases: (1) we make two parallel passes with the base model $M$ to generate constrained and unconstrained predictions ($\hat{y_c}$ and $\hat{y_u}$) and (2) we send $(x, \hat{y_u}, \hat{y_c})$ to the boosted model $M_b$ to make a final prediction $\hat{y}$. This prediction can be made with or without constrained decoding.

In the following sections, we apply \method{} to the cIE task by curating the data and modeling to fit its needs. We emphasize that this paradigm can be used for other structured tasks, with adaptations of the data and modeling. Also note that we use only one step of boosting in our pipeline, although in principle there is nothing that restricts this pipeline to one iteration only. For our setting, we found one step to be sufficient, but for other applications, it is possible to explore multiple iterations of the same algorithm.

\section{Application to information extraction}
\label{sec:application_ie}

To assess \method{}, we apply it to the cIE task and refer to the resulting boosted model as \model{}.

\subsection{Data}
\label{sec:method_data}
To train the base model, we need data that is exhaustive, i.e. the input is fully expressible under constraints. In other words, there should be no facts in the text that we cannot express with entities and relations from the KB. We need this to train a model that is exhaustive. For cIE, this means that the base model should extract all the facts present in the text, regardless of constraints (i.e., perform open information extraction). If this was not the case, the model will likely learn to drop random triplets, and hence the performance would drop. 
For the boosted model, we can simulate the setting in which some samples express entities in the text that do not exist in the KB. For a fraction of the data we randomly remove some entities from the KB making it impossible for a base model to generate them in the constrained setting. We also remove these entities from the target triplet set by removing each triplet that contains the entity in question. By providing these samples during training, we let the model learn what happens when the entity in text is not present in the KB and hopefully bring it closer to generating the correct output.

By curating the data for the boosted model this way, we also prevent the boosted model from learning what is present in the KB, as this changes for every data point. Instead, the boosted model is forced to learn patterns in the constrained and unconstrained outputs from the base model and rely on input information. This makes the model more flexible if KB is changing over time.
Data generated in this way also has some samples for which no triplets are extractable (i.e. they are not present in the KB). As a result, the boosted model is trained to produce an empty set for some samples, which might not be the case for the base model. This does not guarantee that the boosted model would be able to do it for the text that has no triplets at all, but from our results, this seems to be the case.

\subsection{Model and inference for cIE}
\label{sec:method_model}

\xhdr{Modeling}
We follow the same setting for modeling as \citet{synthie}. Both base and boosted models are based on FlanT5 \cite{chung2022scalinginstructionfinetunedlanguagemodels}, and are trained to autoregressively generate a linearized sequence of the corresponding triplet set $y$ when prompted with the input text $x$. Training is done by maximizing the target sequence's conditional log-likelihood with teacher forcing \cite{10.5555/3104482.3104610} and cross-entropy loss. We also use dropout \cite{JMLR:v15:srivastava14a} and label smoothing \cite{7780677}.

\xhdr{Output linearization}
We represent triplets as model-compatible sequences using delimiters: [s], [r], [o], and [e] mark the subject, relation, and object, and the end of each triplet. We concatenate the triplets in the order they appear in the text to form the final sequence.

\xhdr{Inference}
Similarly to \citet{synthie}, we use constrained beam decoding during inference time. Valid prefixes that follow both linearization and KB constraints are dynamically generated.

\subsection{DPO finetuning}
\label{sec:dpo}

As we currently do not have access to a well-aligned dataset for cIE that is made on real-world data (see \Secref{sec:eval_setup}), the process of training base and boosted models is done with synthetic data that might not highly resemble natural text. As a consequence, this might hinder the performance of our model in the wild. In an attempt to overcome this, we propose to tune the model with Direct Preference Optimization (DPO) \cite{rafailov2023direct}, using data more similar to the real-world one. 

DPO is a reward-free method for aligning language models with human preferences by directly optimizing for preferred outputs over less preferred alternatives. In our case, we use DPO to adapt the model toward generating more accurate and faithful structured outputs on real-world text.

For DPO finetuning, we use around 600 samples from the REBEL dataset \cite{rebel}, and an additional 100 samples for validation. We chose this dataset because it was crafted from the real Wikipedia text, although only by collecting text from the first paragraphs of Wikipedia articles. That means that it still differs from randomly crawled Wikipedia text. To identify samples the most similar to the real text, we train a RoBERTa classifier \cite{liu2019robertarobustlyoptimizedbert} that can distinguish real text (random text from Wikipedia articles, not limited to the abstracts) from \synthiedata\ text (for more details about the classifier, see \Appref{appendix:roberta_classifier}). We use this classifier to pick the samples with the highest probability of being real Wikipedia text. Since cIE model trained on this data, GenIE \cite{genie}, is not exhaustive, and SynthIE does not perform well outside of the training distribution, we decide to use a large language model to choose samples with a fitting target from one of these two models (if any of the two can produce it). For each text sample, to collect ranked candidate triplet sets, we run GenIE and SynthIE in constrained manner. We then let GPT-4\footnote{We use {\fontfamily{pcr}\selectfont gpt-4-0613\xspace} version of GPT-4.} decide which one of the two options is better and use this information for ranking. If none of the options are good, we discard the sample. This way, we collect the data that (1) resembles real text more and (2) for which we have a quality solution from one of the existing models. This procedure allows the model to learn a preference signal aligned with real-world patterns, without requiring gold-standard annotations. We note that this might systematically discard harder samples, but is a good starting point to attempt to generalize to a different data distribution.

\section{Evaluation setup}
\label{sec:eval_setup}

\begin{table*}[t]
\centering
\resizebox{\textwidth}{!}{
\setlength{\tabcolsep}{5pt}
\begin{tabular}{lccc|ccc|ccc}
\toprule
 & \multicolumn{3}{c}{\textbf{Overall}} & \multicolumn{3}{c}{\textbf{Removed}} & \multicolumn{3}{c}{\textbf{Same}} \\
 &                          Precision &                          Recall &                         F1 &                          Precision &                          Recall &                         F1 &                          Precision &                          Recall &                         F1 \\
\midrule
\midrule
\textit{\textbf{Micro}}                                &                            &                            &                            &                            &                            &                            &                            &                            &                            \\
\midrule
\hspace{4mm} BoostIE (constrained)          &  57.23 {\scriptsize± 0.79} &  \textbf{48.24 {\scriptsize± 0.63}} &  \textbf{52.35 {\scriptsize± 0.63}} &  38.69 {\scriptsize± 1.84} &  \textbf{45.99 {\scriptsize± 1.79}} &  42.02 {\scriptsize± 1.64} &  63.91 {\scriptsize± 0.94} &  \textbf{48.79 {\scriptsize± 1.03}} &  55.33 {\scriptsize± 0.96} \\
\hspace{4mm} BoostIE (unconstrained)        &  54.72 {\scriptsize± 0.89} &  46.31 {\scriptsize± 0.73} &  50.16 {\scriptsize± 0.73} &  31.62 {\scriptsize± 1.62} &  43.76 {\scriptsize± 1.82} &  36.71 {\scriptsize± 1.57} &  \textbf{64.86 {\scriptsize± 0.91}} &  46.93 {\scriptsize± 1.01} &  54.45 {\scriptsize± 0.95} \\
\hspace{4mm} BoostIE + DPO (constrained)   &  \textbf{59.45 {\scriptsize± 0.65}} &  46.65 {\scriptsize± 0.65} &  52.28 {\scriptsize± 0.59} &  \textbf{43.03 {\scriptsize± 1.82}} &  43.90 {\scriptsize± 1.68} &  \textbf{43.46 {\scriptsize± 1.55}} &  64.82 {\scriptsize± 0.94} &  47.31 {\scriptsize± 0.96} &  \textbf{54.70 {\scriptsize± 0.91}} \\
\hspace{4mm} BoostIE + DPO (unconstrained) &  56.39 {\scriptsize± 0.82} &  44.81 {\scriptsize± 0.73} &  49.94 {\scriptsize± 0.72} &  35.41 {\scriptsize± 1.80} &  41.94 {\scriptsize± 1.80} &  38.39 {\scriptsize± 1.65} &  64.58 {\scriptsize± 1.02} &  45.50 {\scriptsize± 0.95} &  53.38 {\scriptsize± 0.93} \\
\hspace{4mm} \relik{} (filtered)      &  22.89 {\scriptsize± 0.57} &  20.80 {\scriptsize± 0.58} &  21.79 {\scriptsize± 0.53} &  17.21 {\scriptsize± 1.01} &  18.37 {\scriptsize± 1.12} &  17.77 {\scriptsize± 0.96} &  24.58 {\scriptsize± 0.66} &  21.43 {\scriptsize± 0.56} &  22.89 {\scriptsize± 0.57} \\
\hspace{4mm} SynthIE 400k (constrained)             &  31.71 {\scriptsize± 0.77} &  39.81 {\scriptsize± 0.68} &  35.30 {\scriptsize± 0.68} &  13.57 {\scriptsize± 0.92} &  40.41 {\scriptsize± 1.85} &  20.31 {\scriptsize± 1.20} &  45.90 {\scriptsize± 1.03} &  39.67 {\scriptsize± 0.83} &  42.56 {\scriptsize± 0.88} \\
\hspace{4mm} SynthIE 400k (unconstrained)           &  33.40 {\scriptsize± 0.81} &  34.83 {\scriptsize± 0.77} &  34.10 {\scriptsize± 0.73} &  15.18 {\scriptsize± 0.98} &  34.54 {\scriptsize± 1.73} &  21.09 {\scriptsize± 1.19} &  45.98 {\scriptsize± 1.17} &  35.00 {\scriptsize± 0.88} &  39.75 {\scriptsize± 0.96} \\

\midrule
\midrule

\textit{\textbf{Macro}}                                &                            &                            &                            &                            &                            &                            &                            &                            &                            \\
\midrule
\hspace{4mm} BoostIE (constrained)          &  58.35 {\scriptsize± 2.46} &  \textbf{46.11 {\scriptsize± 1.02}} &  \textbf{48.81 {\scriptsize± 1.39}} &  37.51 {\scriptsize± 2.55} &  \textbf{39.47 {\scriptsize± 3.11}} &  36.01 {\scriptsize± 2.20} &  \textbf{61.96 {\scriptsize± 3.30}} &  \textbf{46.26 {\scriptsize± 1.62}} &  \textbf{50.28 {\scriptsize± 1.95}} \\
\hspace{4mm} BoostIE (unconstrained)       &  43.81 {\scriptsize± 1.78} &  35.78 {\scriptsize± 0.97} &  37.34 {\scriptsize± 1.20} &  26.69 {\scriptsize± 2.14} &  32.59 {\scriptsize± 3.04} &  27.24 {\scriptsize± 2.01} &  53.21 {\scriptsize± 3.11} &  38.12 {\scriptsize± 1.23} &  42.31 {\scriptsize± 1.63} \\
\hspace{4mm} BoostIE + DPO (constrained)   &  \textbf{59.09 {\scriptsize± 2.50}} &  44.74 {\scriptsize± 1.10} &  48.29 {\scriptsize± 1.50} &  \textbf{39.32 {\scriptsize± 2.93}} &  38.34 {\scriptsize± 2.70} &  \textbf{36.55 {\scriptsize± 2.12}} &  61.58 {\scriptsize± 3.21} &  45.00 {\scriptsize± 1.58} &  49.27 {\scriptsize± 1.87} \\
\hspace{4mm} BoostIE + DPO (unconstrained) &  42.89 {\scriptsize± 1.75} &  32.91 {\scriptsize± 0.97} &  35.23 {\scriptsize± 1.07} &  28.33 {\scriptsize± 2.18} &  31.59 {\scriptsize± 2.51} &  27.85 {\scriptsize± 1.71} &  50.84 {\scriptsize± 3.18} &  35.54 {\scriptsize± 1.29} &  39.73 {\scriptsize± 1.66} \\
\hspace{4mm} \relik{} (filtered)     &  17.22 {\scriptsize± 0.99} &  12.81 {\scriptsize± 0.54} &  12.92 {\scriptsize± 0.56} &  11.63 {\scriptsize± 1.38} &  11.96 {\scriptsize± 0.80} &  10.59 {\scriptsize± 0.75} &  17.20 {\scriptsize± 1.36} &  13.14 {\scriptsize± 0.53} &  13.18 {\scriptsize± 0.66} \\
\hspace{4mm} SynthIE 400k (constrained)           &  40.29 {\scriptsize± 1.60} &  38.77 {\scriptsize± 0.95} &  36.25 {\scriptsize± 1.12} &  21.22 {\scriptsize± 1.87} &  31.72 {\scriptsize± 3.54} &  22.67 {\scriptsize± 2.11} &  47.97 {\scriptsize± 2.15} &  38.26 {\scriptsize± 0.73} &  39.68 {\scriptsize± 1.21} \\
\hspace{4mm} SynthIE 400k (unconstrained)        &  35.58 {\scriptsize± 1.28} &  33.76 {\scriptsize± 1.16} &  32.28 {\scriptsize± 1.05} &  16.94 {\scriptsize± 1.39} &  27.10 {\scriptsize± 3.41} &  18.88 {\scriptsize± 1.71} &  46.26 {\scriptsize± 2.53} &  34.04 {\scriptsize± 0.86} &  37.04 {\scriptsize± 1.34} \\

\bottomrule

\end{tabular}
}
\caption{Results on \synthiedata{} dataset: Overall - whole test set, Removed - test samples with removed random entities (and triplets) from the target and KB, Same - test samples without entity removal. For \model{}, constrained and unconstrained refers to the final boosted model mode of operation. We report both micro and macro results, with 95\% CI. Best results are in bold.}
\label{tab:main}
\end{table*}

\xhdr{Knowledge base}
We use the subset of Wikidata \cite{10.1145/2187980.2188242}, using entities that are connected to English Wikipedia pages and relations that appear at least once in the REBEL training dataset. Our catalogue consists of 2.6M entities and 888 relations. For the unique representation of each entity, we use its English Wikipedia title. For relations, we use their label in Wikidata.

\xhdr{Data}
For training the base model $M$, we use a subset of 300K samples from the train split of \synthiedata\ used for training SynthIE models (see \Appref{appendix:synthie_data} for details). This is a synthetic dataset and we use it because there is no real-text dataset with inputs that are fully expressible under KB constraints (see \Secref{sec:method_data} for a further explanation of this requirement). The boosted model $M_b$ was trained on an additional 100K samples from the same dataset. We also train a SynthIE model on all 400k samples for a fair comparison.
100K samples used for training the boosted model have been altered as explained in \Secref{sec:method_data}, and 40\% of the randomly chosen samples have been altered. For each altered sample, up to three entities were removed, uniformly. The validation and test data were crafted in the same way, each being a subset of the corresponding \synthiedata{} of 10K samples. \synthiedata{} is imperfect as it does not have the same properties as the real-world text, but can demonstrate the abilities of this training technique effectively.

\xhdr{Baselines}
To isolate the effects of this training technique, we compare \model{} with the SynthIE model of the same size, trained on the same 400K samples used in the \model{} pipeline (without alterations). We also compare our method with \relik\ cIE model of similar size,\footnote{We use ``relik-ie/relik-cie-large'', see \url{https://huggingface.co/relik-ie/relik-cie-large}} as this is the state-of-the-art model right now. We provide results with and without using DPO after initial training. For more details about the baselines, see \Appref{appendix:baselines}.

\xhdr{Metrics and implementation detail}
We evaluate the performance in terms of micro and macro precision, recall, and F1 score. All results are reported with 95\% confidence intervals constructed from 50 bootstrap samples. For more details on evaluation metrics, see \Appref{appendix:metrics}. For details on implementation, see \Appref{appendix:implementation}.

\section{Results}

\subsection{Evaluation on \synthiedata}
\label{sec:eval_synthie}

\xhdr{Performance evaluation}
We first evaluate our method on in-distribution data. We use the metrics mentioned in \Secref{sec:eval_setup} on the random subset of 10K samples from the test split of \synthiedata. We evaluate it on non-edited, as well as \synthiedata{} with entities randomly removed from the KB, as described in \Secref{sec:method_data}. We report results in \Tabref{tab:main}.

\relik\ does not perform on \synthiedata{} nearly as well as SynthIE and \model{}. This is expected, as it was not trained on this data, and \synthiedata{} has a different distribution from REBEL on which \relik{} was trained.

Second, we notice that all the models perform worse for the samples where some entities are randomly removed from the KB. This is in line with our expectations, especially for SynthIE, as it was trained to extract exhaustively, and cannot handle instances where this is not possible. Precision is more affected by this modification of the data. Micro-recall stays almost the same, while macro-recall drops much less than precision. This happens because the models tend to output wrong triplets either related to the removed entity (in unconstrained mode) or related to a similarly named entity (in constrained mode). Triplets related to the correct entities present in the graph mostly stay in the output, maintaining the recall relatively high.

For \model{} models, there is a noticeable improvement both for edited samples with removed entities and for the non-edited ones. The improvement is visible for both micro and macro scores. We suspect that this happens because by using \method{} (1) we are implicitly including the information about the presence or lack of an entity in the KB and (2) we include the information about errors SynthIE, which is used as a base model, makes regardless of the KB. Examples of the latter can be wrong disambiguation of certain entities in the KB, or the less adequate relations for the scenario (for instance, using ``location'' instead of ``located in or next to body of water'' for text expressing an entity ``Niagara'' being located in the ``Lake Ontario'').

The difference in scores for constrained and unconstrained settings is higher for \model{} models than for SynthIE. 
This happens because, unlike SynthIE which tends to generate triplets with wrong entities when something is not present in the KB, \model{} is able to recognize this setting. This is expected, as \method{} used for training \model{} models specifically addresses this issue. 
We speculate that, in the case of \model{}, constrained decoding helps filter out triplets with missing entities rather than causing the model to generate triplets with incorrect ones. In other words, \model{} assigns a higher probability to the output that does not include entities missing from the KB. For macro scores, the difference is present for both original and edited samples. This likely means that \model{} detects some systematic errors that happen for rare relations when using SynthIE.

Finally, the usage of DPO does not result in significant improvements over the standard \model{} model on this data. This is expected given its use was aimed at improving real-data performance (see \Secref{sec:eval_real} for evaluation on natural text). Still, the absence of performance degradation, for both micro and macro socres, is a positive sign. 

\begin{figure}[t]
    \centering    \hspace{-3pt}\includegraphics[width=\columnwidth]{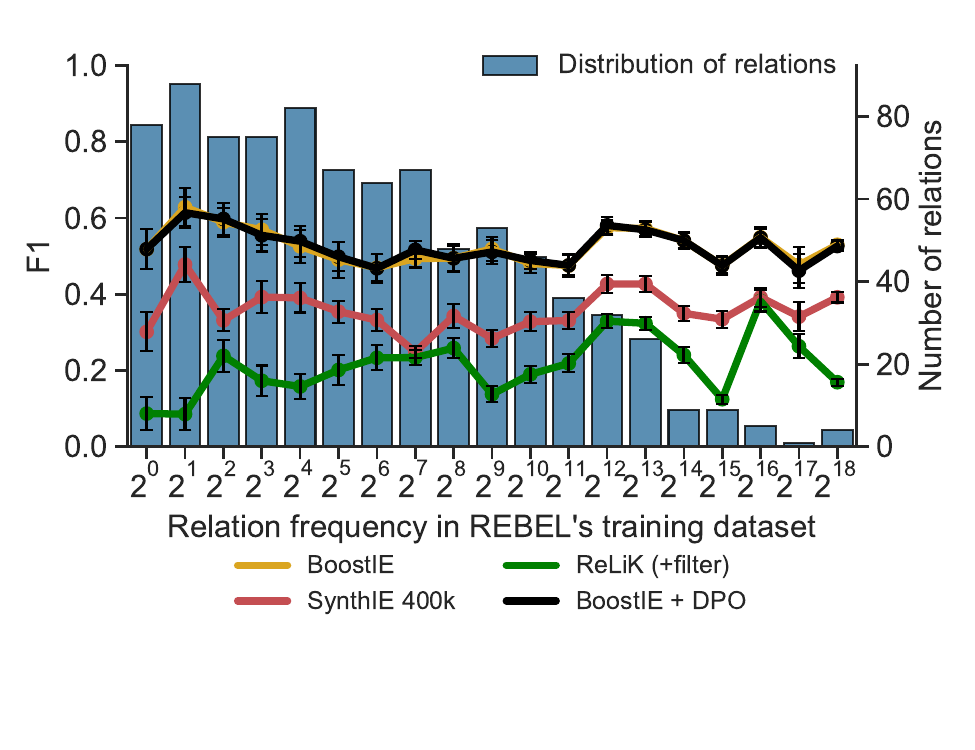}
    \vspace{-5mm}
    \caption{\textbf{Impact of the relation frequency.} Relations are bucketed based on their frequency; bucket $2^i$ contains relations occurring between $2^{i}$ and $2^{i+1}$ times. The histogram shows the number of relations per-bucket. The line plots depict the per bucket F1 scores evaluated on \synthiedata{} test dataset with confidence intervals constructed by bootstrapping.}
    \label{fig:bucket_plot}
\end{figure}
\xhdr{Performance by relation frequency}
As mentioned earlier, relations expressed in the natural text are imbalanced: there is a small number of relations that are present very often and a large number that are rare. Training on real data can lead to bad performance on those rare relations, which would be masked by the overwhelming presence of common relations. \synthiedata{} was constructed with this in mind. To verify that our method does not compromise the performance on rare relations, as well as to evaluate the performance of \relik{} in this light, we mimic the experiment by \citet{synthie} and bucket relations by their frequency in REBEL training set which follows the natural distribution of relations. We report results in \Figref{fig:bucket_plot}. \relik{} performs worse for rare relations. This is expected, as parts of their pipeline were trained with real-world data. When it comes to \model{} models, they perform consistently better than SynthIE for all relation buckets and maintain stable performance over rare and common relations.

\subsection{Evaluation on natural text}
\label{sec:eval_real}
\begin{table}[htp]
    \centering
    \resizebox{\columnwidth}{!}{
    \setlength{\tabcolsep}{5pt}
\begin{tabular}{llll}
\toprule
\textbf{\textit{Micro}}                        &           \textbf{Precision}                  &              \textbf{Recall}               &           \textbf{F1}                  \\
\midrule
\hspace{4mm}\model{} + DPO &  \textbf{48.89 {\scriptsize± 18.16}} &  \textbf{27.76 {\scriptsize± 11.78}} &  \textbf{34.93 {\scriptsize± 11.97}} \\
\hspace{4mm}\model{}     &  22.68 {\scriptsize± 11.85} &  17.38 {\scriptsize± 11.02} &  19.33 {\scriptsize± 10.12} \\
\hspace{4mm}\relik{}       &  25.88 {\scriptsize± 18.38} &  22.58 {\scriptsize± 15.33} &  23.99 {\scriptsize± 16.38} \\
\hspace{4mm}SynthIE 400k     &    6.74 {\scriptsize± 3.91} &  13.34 {\scriptsize± 10.19} &    8.76 {\scriptsize± 4.96} \\
\midrule
\midrule
\textbf{\textit{Macro}}                        &                             &                             &                             \\
\hspace{4mm}\model{} + DPO &  \textbf{23.87 {\scriptsize± 4.92}} &  \textbf{20.85 {\scriptsize± 3.91}} &  \textbf{20.87 {\scriptsize± 4.13}} \\
\hspace{4mm}\model{}     &  15.50 {\scriptsize± 3.20} &  13.62 {\scriptsize± 2.88} &  13.49 {\scriptsize± 2.77} \\ 
\hspace{4mm}\relik{}       &   9.52 {\scriptsize± 0.95} &   9.55 {\scriptsize± 2.84} &   8.33 {\scriptsize± 1.35} \\
\hspace{4mm}SynthIE 400k     &   5.38 {\scriptsize± 1.95} &   6.05 {\scriptsize± 2.66} &   5.35 {\scriptsize± 2.11} \\
\bottomrule

\end{tabular}
}
\caption{Human evaluation on Wikipedia text. The best results are bolded. Results are reported with 95\% CI.}
\label{tab:human_eval}
\end{table}

To better understand the performance of both our \model{} models, as well as \relik{} and SynthIE out of distribution, we manually annotate 50 random samples from Wikipedia text (see \Appref{appendix:wikipedia_collection} for data collection process). During this process, each Wikipedia text sample is assigned a ground truth triplet set (see \Appref{appendix:human_eval} for annotation process details). We then compare all models on this data. We run both \model{} and SynthIE with constrained decoding, as evaluations on \synthiedata{} suggest this improves performance. Results are presented in \Tabref{tab:human_eval}. We note that Wikipedia text does not fully reflect the real-world text, as it is a highly structured and factual text. Nonetheless, it is a good starting point for evaluating cIE models.

From \Tabref{tab:human_eval}, SynthIE performs the worst. This matches our expectations as SynthIE was trained on the data that has little resemblance to the Wikipedia text. Next, \model{} performs slightly worse than \relik{} in terms of micro metrics, but a little better in terms of macro metrics. We see this as a good sign. \model{} manages to be on par with \relik{} without the need for a separate retrieval model. We suspect that the reason why \model{} is better in macro metrics is because \relik{} was trained on the REBEL dataset, which has heavy-tailed distribution of relations. Finally, \model{} with DPO performs by far the best over both micro and macro metrics. This highlights the importance of the DPO step, and the potential it has to adapt the language model to the a differently distributed data.


\subsection{Error analysis}
\label{sec:error_analysis}

To further examine what kind of errors previous and our method make, we collect 50 random samples of text from Wikipedia (see \Appref{appendix:wikipedia_collection} for the data collection process). We compare SynthIE, \relik{} and \model{} with and without the DPO step. By manual inspection, we identify five types of errors:

(i) \textbf{Unexhaustive triplets:} triplet set does not include some correct triplets

(ii) \textbf{Incorrect related triplets:} triplet set includes some incorrect triplets about correct entities

(iii) \textbf{Misclassified entity:} entities in the triplets are wrongly identified as similarly named ones

(iv) \textbf{Unrelated triplets:} triplet set includes triplets unrelated to the text or entities in the text

(v) \textbf{Entity-centered triplets:} triplets in the triplet set are centered around one entity

Some errors can happen at the same time, e.g. there can be a triplet set that is both unexhaustive and contains unrelated triplets. We annotate the chosen sample and report the results in \Tabref{tab:error_analysis_full}. 

From the results, it is clear that SynthIE struggles with the Wikipedia data in multiple ways. Most of the samples contain at least some unrelated triplets (60\%). We also notice that it has the highest percentage of samples with misclassified entities (9\%). Both of these errors stem from the constrained decoding issues -- when the entity is not present in the KB but is expressed in the text, SynthIE tends to produce triplets with similarly-named entities (misclassified) or even completely unrelated ones. This is confirmed by the \model{} results, as both of these problems are largely mitigated for \model{}.

SynthIE also produces the highest percentage of samples with triplet sets centered around one entity (16\%). We notice that \model{} without DPO has similar issues (11\%). We believe this error comes from a bad distribution of triplet sets in the \synthiedata{} used for training both of these models. \relik{} and \model{} with DPO which were either trained with different data (REBEL), or exposed to it through DPO, suffer from this suffer from this to a much lesser degree (0\% and 6\% respectively).

Among all error types, unexhaustive generations exhibit the least variance across the four models.
Despite intentionally training SynthIE and \model{} models on an exhaustive dataset, on the real text, they fall short similarly to \relik{} trained on an unexhaustive dataset (REBEL). 
We suspect that the limited performance of \model{} without DPO might be due to a significant mismatch between the training data distribution and the real-world text. In the case of \model{} with DPO, although the data used during fine-tuning more closely resembles Wikipedia text, it includes some outputs from GenIE, which is trained on REBEL. We expect that some of these outputs are not exhaustive. This likely contributed to the persistence of unexhaustive generations. In \Appref{appendix:additional_analyses}, we provide a few additional analyses of common cIE approaches, setting the stage for further research in this area.

\begin{table}[htp]
    \centering
    \resizebox{\columnwidth}{!}{
    \setlength{\tabcolsep}{5pt}
    \begin{tabular}{lcccc}
    \toprule
         & \textbf{SynthIE} & \textbf{\relik{}} & \textbf{\model{}} & \textbf{\model{} + DPO} \\
    \midrule
    \midrule
    Unexhaustive                                                                       & 0.33 {\scriptsize± 0.09} &  0.38 {\scriptsize± 0.11} & 0.32 {\scriptsize± 0.11} & 0.38 {\scriptsize± 0.10} \\
    Incorrect related                     & 0.36 {\scriptsize± 0.11}   & 0.28 {\scriptsize± 0.10}                           & 0.26 {\scriptsize± 0.11} & 0.12 {\scriptsize± 0.09} \\
    Misclassified entity     & 0.09 {\scriptsize± 0.07}   &     0.04 {\scriptsize± 0.04}                                      & 0.00 {\scriptsize± 0.00} & 0.00 {\scriptsize± 0.00} \\
    Unrelated     & 0.60 {\scriptsize± 0.11}        & 0.14 {\scriptsize± 0.09}                    & 0.28 {\scriptsize± 0.10} & 0.08 {\scriptsize± 0.06} \\
    Entity-centered & 0.16 {\scriptsize± 0.08} & 0.00 {\scriptsize± 0.00} & 0.11 {\scriptsize± 0.07} & 0.06 {\scriptsize± 0.05} \\
    \bottomrule
    \end{tabular}
    }

    \caption{Error analysis on Wikipedia text samples. Numbers represent fraction of samples with the given type of error. Result are shown with 95\% CI.}
    \label{tab:error_analysis_full}
\end{table}




\section{Related work}

\subsection{Closed information extraction}

Older cIE methods usually rely on the combination of entity recognition \cite{tjong-kim-sang-2002-introduction} and linking \cite{10.1145/1458082.1458150} with relation extraction \cite{relation_extraction} to obtain the set of triplets constrained to the KB. However, these methods often have problems with error propagation due to their architecture \cite{mesquita-etal-2019-knowledgenet, trisedya-etal-2019-neural}. A newer approach that combines entity linking and relation extraction is proposed by \citet{relik}. In recent years, however, autoregressive methods have dominated the field. For the cIE task, this was first introduced by \citet{genie}. \citet{genie} also introduced the usage of constrained decoding for this task. The same approach was adopted by \citet{synthie} and \citet{webie}.

Another line of research in this field relies on building a good training dataset for the cIE task. \citet{rebel} introduced REBEL, a dataset of fact triplets constructed using distant supervision. Similarly, \citet{trisedya-etal-2019-neural} introduce WikiNER, a dataset that is also made using distant supervision, but augmented with co-reference resolution and dictionary-based paraphrase detection. More recently, \citet{webie} presented WebIE, a multilingual distant-supervision dataset, with the introduction of some human-annotated samples as well. \citet{synthie} synthetically generated their data specifically having distributional (i.e. relational frequency issue) and exhaustiveness issues in mind.

The emergence of LLMs raises the question of their ability to perform this task. As shown by \citet{synthie}, LLMs struggle with tasks that require structured output. For cIE, they also have no knowledge of the KB. \citet{sgcd} attempt to overcome this issue by combining an LLM with constrained decoding, but their evaluation on synthetic data limits broader conclusions.

\subsection{Constrained decoding}

Structured NLP tasks require the output to be in a certain form. To overcome this, different forms of constrained decoding have been proposed. \citet{decao2021autoregressiveentityretrieval} address the entity-disambiguation constraints by generating a prefix trie at the decoding time, forcing output to be valid entities from the KB. \citet{geng2024grammarconstraineddecodingstructurednlp} introduce grammar-constrained decoding, focusing on generalizing the constrained decoding to a wider variety of tasks. \citet{park2024grammaraligneddecoding} introduce grammar-aligned decoding, which aims to correct the conditional probability of the LLM's distribution conditioned on the given grammar constraint. \citet{koo2024automatabasedconstraintslanguagemodel} propose a method that addresses downsides of constrained decoding related to the tokenization issues by using automata-based constraints. \citet{beurerkellner2024guidingllmsrightway} propose a method that speeds up the constrained decoding that works in a subword-aligned fashion.

    


\section{Discussion}

\subsection{Implications for cIE}

Despite numerous efforts through years to solve cIE, current approaches struggle with performance on the real data, as well as adaptability to different KBs. Our method could be a step closer to an efficient and high-performing system that overcomes these issues. Our experiments show that \model{} (\method{} applied to cIE) improves the performance of constrained decoding, which is often used for cIE systems. Additionally, \model{} does not directly learn what is present in the KB, which is the case for most current approaches, making it more adaptable to changes in the KB. Our experiments on \synthiedata{} also show that our method maintains a good performance over rare relations, while the evaluation on real Wikipedia data indicates that \model{} is better at generalizing to out-of-distribution data. This seems to be the case especially when using DPO with data that resembles the target distribution. With that in mind, along with the fact that DPO does not degrade performance on the original data distribution, we draw attention that this can be used as an unexpensive way to improve the overall performance of the model. In an ideal scenario, our base model would be trained on an exhaustive dataset with more realistic text. This is not trivial to collect, so finetuning with DPO and a smaller finetuning dataset can be a good way to overcome this limitation.


\subsection{Implications for other tasks}

Although our present evaluation has focused on the benefits of \method{} for closed information extraction, nothing about the method is inherently restricted to this task. A similar pipeline can be exploited for a wide range of structured NLP tasks, including tagging, parsing, code generation, JSON generation, and many more.
We leave the evaluation of \method{} on such other tasks for future work and hope that researchers and developers will benefit from \method{} in practice.



\section*{Limitations}

\xhdr{Entity surface form variations}
Our current pipeline might have issues with entities that are presented in the text in a very different way than in the knowledge base (e.g. as acronyms or aliases). Since our model has no external knowledge, it cannot disambiguate between these cases \vs an entity that is present in the text but not in the KB. This is also something that we cannot expect from a small, specialized, model to know on its own, as it does not have broad knowledge of the external world. This is possibly an area where LLMs would excel.

\xhdr{Inference speed}
Although we are using small language models for this task and we consider our approach to be scalable, inference requires three runs of a model (constrained and unconstrained base model run, and the run of the boosted model). This is less efficient than SynthIE or similar models, but is still faster and cheaper than running an LLM. Also note that the constrained and unconstrained run of the base model can be parallelized.

\xhdr{Training dataset}
The dataset we used for training does not resemble real data, and has other distributional issues. One particular case of such issue is the distribution of entities in the triplet sets. Due to the way \synthiedata{} was generated, most of the triplets in triplet sets are centered around one or two entities. Real data often describes many more entities in a few sentences. Because of this, both SynthIE and \model{} have troubles with text that expresses triplets about many entities in a single sentence or paragraph. This can be solved by different sampling of triplet sets when generating synthetic data for training, focusing on introducing variety of entities into them.

\xhdr{Real-world data performance}
While \model{} improved the overall performance on the sampled Wikipedia text, it is still far from perfect. Additionally, Wikipedia does not fully reflect the performance of our model in the wild, as it is still a very factual and structured text. In future work, it would make sense to perform a further evaluation on the real text, as it might help identify other failure modes.


\section*{Acknowledgments}
We would like to thank Ivan Zakazov, Alexander Sharipov, Lorenzo Drudi, Kamel Charaf, Haolong Li and Saibo Geng for helping with the human evaluation. We also thank Yiyang Feng for help with the initial exploration. West’s lab is partly supported by grants from the Swiss National Science Foundation (200021\_185043 and 211379), Swiss Data Science Center (P22\_-08), H2020 (952215), Microsoft Swiss JRC, and Google, and by generous gifts from Facebook, Google, and Microsoft.

\bibliography{main}

\appendix

\section{Collection of Wikipedia text}
\label{appendix:wikipedia_collection}
We collect Wikipedia text using Wikipedia API, by randomly taking Wikipedia articles and extracting 1 chunk of text that has at most 4 sentences per each article. All the sentences have to be part of the same paragraph (i.e. we are not keeping chunks that contain ``\textbackslash{}n'' in them).

\section{\synthiedata}
\label{appendix:synthie_data}

\synthiedata{} is a fully synthetic dataset introduced by \citet{synthie}. It was used for training the range of SynthIE models. The dataset consists of around 1.8M training data samples, 10K validation, and 50K test samples generated by the now discontinued OpenAI model, {\fontfamily{pcr}\selectfont code\hyp davinci\hyp 002\xspace}. The data was synthetically made, starting from sampling triplet sets. Triplet sets are generated by a biased random walk on a subset of the Wikidata knowledge graph \cite{10.1145/2187980.2188242}. Text that corresponds to these triplets was then generated by an LLM. Each text sample was generated by providing a triplet set and asking the LLM to write the text that only expresses those triplets. This way, an exhaustive, high-quality data was made. The main disadvantage of this dataset is the fact that the text does not resemble real text, as it is very clean and does not contain a lot of details.

\section{Baselines}
\label{appendix:baselines}

\xhdr{GenIE}
\citet{genie} introduce GenIE, an end-to-end autoregressive langauge model that does cIE, based on BART \cite{lewis-etal-2020-bart}. This model was trained on REBEL, the dataset made with distant supervision on Wikipedia abstracts. The method also employs constrained decoding. Given all of this, the model has issues that stem from constrained decoding, bad alignment between triplets and text in the data, as well as bad distribution of relations in the training set. We do not compare against GenIE as it was already shown by \citet{synthie} that it performs worse than SynthIE. We use it to generate DPO data (see \Secref{sec:dpo}).

\xhdr{SynthIE}
As a part of efforts to mitigate some of the issues raised by GenIE, \citet{synthie} introduce SynthIE. This is a model trained on synthetic data, \synthiedata{}, that has better alignment between text and triplets, as well as better distribution of relations in the training set. However, SynthIE still uses constrained decoding, and the synthetic data it was trained on does not resemble real data, which causes issues when the model is used in practical settings.

\xhdr{\relik}
Differently from SynthIE and our \model{} models, \relik{} \cite{relik} utilizes a retriever-reader architecture to solve cIE task. The retriever module encodes the input text and retrieves the most relevant entities and relations from the KB. Then, the reader module takes as input the text and each retrieved entity or relation separately and maps them to a specific span of the text. The modules for cIE were trained on REBEL dataset \cite{rebel}, raising a concern that this model might exhibit the issues with rare-relation performance.

\section{Metrics}
\label{appendix:metrics}

We evaluate performance using standard precision, recall, and F1 metrics across all settings. A predicted fact is considered correct only if the relation and both associated entities are correct. Formally, let the set of predicted triples for a document $d \in \mathcal{D}$ be denoted as $P_d$, and the corresponding set of gold triples as $G_d$. Then, the micro-averaged precision and recall are defined as follows: 

\begin{align} 
\text{micro-precision} &= \sum_{d \in \mathcal{D}} |P_d \cap G_d| \bigg/ \sum_{d \in \mathcal{D}} |P_d|, 
\end{align}

and

\begin{align} 
\text{micro-recall} &= \sum_{d \in \mathcal{D}} |P_d \cap G_d| \bigg/ \sum_{d \in \mathcal{D}} |G_d|. 
\end{align}

Micro scores provide a useful aggregate view of model performance, especially in terms of overall accuracy. However, they can obscure disparities in datasets with class imbalance—for instance, when certain entities or relations appear far more frequently in both training and test data. This is because micro-averaging gives equal weight to each instance, whereas macro-averaging assigns equal weight to each class. To account for such imbalances, we also report macro-averaged scores.

Let $P_d^{(r)}$ and $G_d^{(r)}$ denote the predicted and gold triples for relation $r \in \mathcal{R}$ in document $d$. Then, macro-precision is defined as: 

\begin{align}
\frac{1}{\mathcal{|R|}} \sum_{r \in \mathcal{R}} \left(; \sum_{d \in \mathcal{D}} |P_d^{(r)} \cap G_d^{(r)}| \bigg/ \sum_{d \in \mathcal{D}} |P_d^{(r)}| \right) , \end{align} 

and macro-recall as:

\begin{align} 
\frac{1}{\mathcal{|R|}} \sum_{r \in \mathcal{R}} \left(; \sum_{d \in \mathcal{D}} |P_d^{(r)} \cap G_d^{(r)}| \bigg/ \sum_{d \in \mathcal{D}} |G_d^{(r)}| \right) . 
\end{align}

\section{Implementation}
\label{appendix:implementation}
As mentioned in \Secref{sec:method_model}, \model\ uses two FlanT5 models. For both models, we use 'google/flan-t5-base' version\footnote{\url{https://huggingface.co/google/flan-t5-base}}, which has $\sim$250M parameters. The models were trained using the Adam optimizer with a learning rate of 3e-4, 0.1 gradient clipping on the Euclidean norm, and a weight decay of 0.05. They were trained with batch size 80, for a maximum of 10K steps. We used a polynomial learning rate scheduler with 1000 warm-up steps and a final learning rate of 3e-05. All the experiments were run on a single NVIDIA Titan X Maxwell 12GB GPU, taking around 24h for the training of the base model, and around 16h for boosted model. The DPO finetuning was done on the same machine with the \textit{trl} library\footnote{\url{https://huggingface.co/docs/trl/en/index}}, using learning rate 5e-5, batch size 2, $\beta$ 0.1 and running it for 5 epochs, taking around 20min to finetune. During inference, we run all our models with 10 beams.

\section{DPO data preprocessing}
\label{appendix:roberta_classifier}

To collect the data for DPO finetuning, we first train a RobERTa classifier that distinguishes \synthiedata{} text from real Wikipedia text. We use the 'roberta-base' model\footnote{\url{https://huggingface.co/FacebookAI/roberta-base}} as the basis for our classifier. To do that, we take 5K samples from the \synthiedata{} training split (labeled as '0') and collected 5K samples of Wikipedia text (labeled as '1') in the way described in \Appref{appendix:wikipedia_collection}. For the validation set, we collect in total of 3K samples in the same way. The classifier achieves an accuracy of 98.27\% on the validation set, highlighting again how different SynthIE data is from the real Wikipedia one.

\section{Human annotations}
\label{appendix:human_eval}

\xhdr{Construction of candidate triplet sets}
We start by randomly choosing 50 samples of Wikipedia text. Since it is not trivial to annotate the text, as the knowledge of a whole KB with more than 2.6M entities and almost 900 relations, we attempt to get as exhaustive set of candidate triplets as possible by combining outputs from multiple models. For that, we use SynthIE, GenIE, \relik, \model, and \model{} with DPO. 

Because these models were trained on different datasets, and have different strengths and disadvantages, by combining all of them, we are hoping to at least have a set of triplet candidates that include all the correct triplets, while also possibly including many incorrect ones. This procedure ensures that our precision estimate is correct, up to human error. For the recall, our estimation will not necessarily be correct, but the ranking of the models will stay the same (as they all might be missing some potential triplets that none of the models generated).

\begin{figure*}[t]
    \centering
    \hspace{-3pt}
    \fbox{\includegraphics[width=\textwidth]{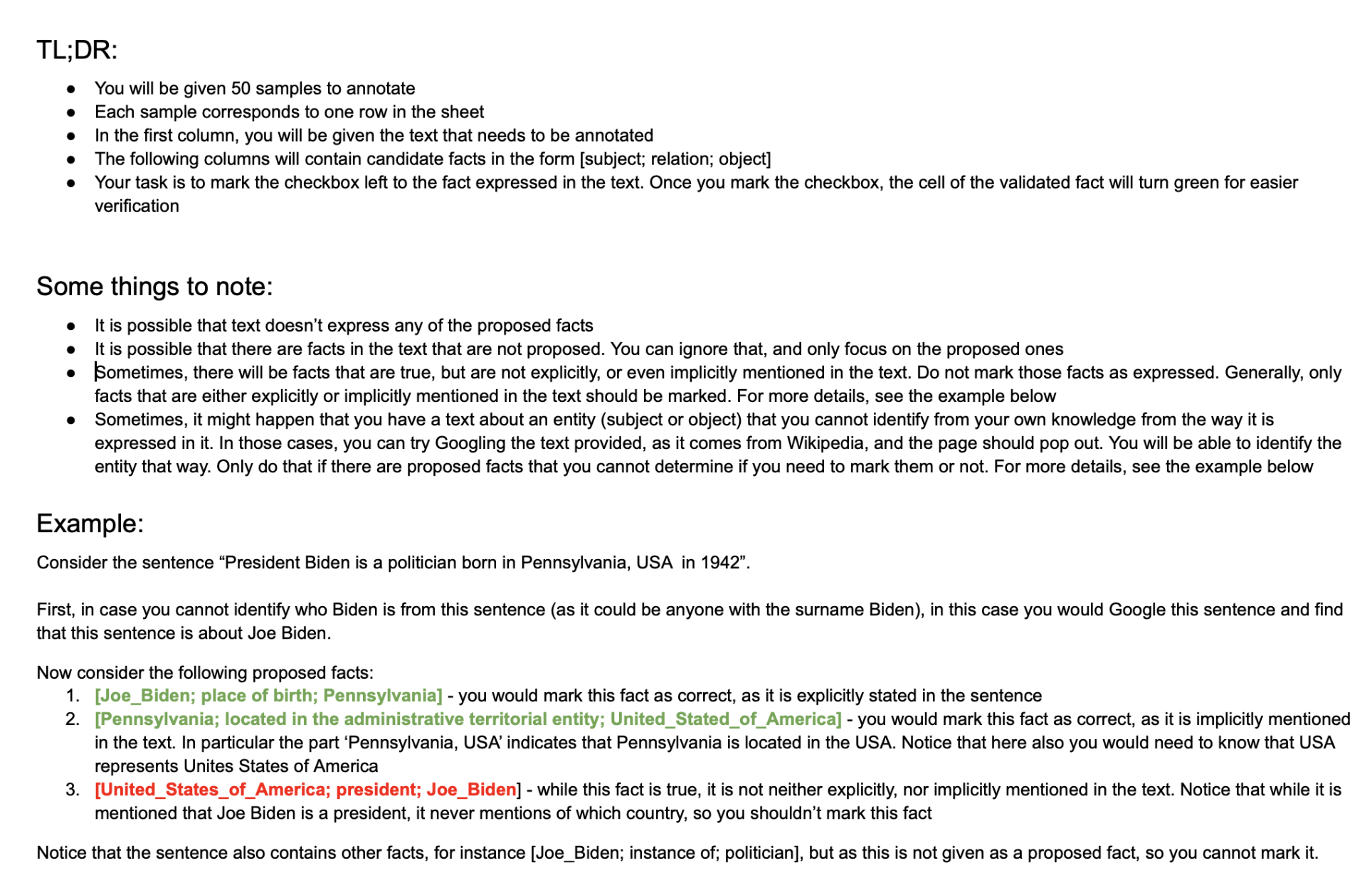}}
    \caption{Human evaluation instructions. Annotators are provided with the sheet with text and candidate triplets, and with the detailed instructions.}
    \label{fig:annotator_instructions}
\end{figure*}

\xhdr{Instructions}
The annotators were given instructions in \Figref{fig:annotator_instructions}.

\xhdr{Annotation task}
To ensure quality results, our annotation was done by 2 Ph.D. students and 4 MSc students. None of them were familiar with our work, avoiding any possible biases. For each annotation sample, the annotator was presented with the text and list of candidate triplets. For each triplet, they had to decide whether the triplet is expressed in the text or not, based on the instructions provided in \Figref{fig:annotator_instructions}. The annotation was done in three stages. First, one Ph.D. student and all MSc students annotated the data, with each contributing to an equal part. Then, the second Ph.D. student annotated all the samples. Finally, one of the paper's authors resolved the conflicts.

\section{Additional analysis of cIE methods}
\label{appendix:additional_analyses}
\begin{table}[]
\centering
    \resizebox{\columnwidth}{!}{
    \setlength{\tabcolsep}{5pt}
\begin{tabular}{@{}lcc
@{}}
\toprule
                   & \textbf{SynthIE (constrained}) & \textbf{SynthIE (unconstrained)} \\ \midrule
\midrule
Unexhaustive                  & 0.33 ± 0.09           & 0.23 ± 0.11             \\
Incorrect related             & 0.36 ± 0.11           & 0.46 ± 0.13             \\
Misclassified entity          & 0.09 ± 0.07           & 0.03 ± 0.03             \\
Unrelated                     & 0.60 ± 0.11           & 0.26 ± 0.11             \\
Entity-centered               & 0.16 ± 0.08           & 0.34 ± 0.11             \\
Entity/relation not in the KB & 0.00 ± 0.00           & 0.72 ± 0.11             \\ \bottomrule
\end{tabular}
}
\caption{Error analysis for SynthIE model in constrained and unconstrained mode on Wikipedia real text.}
\label{tab:synthie_constrained_unconstrained_analysis}
\end{table}
\begin{table*}[]
\centering
\resizebox{\textwidth}{!}{
\setlength{\tabcolsep}{5pt}
\begin{tabular}{@{}p{0.45\textwidth}|p{0.45\textwidth}|p{0.45\textwidth}@{}}
\toprule
\textbf{text}                                                                                                                                                                                                                                                                                                                                                                                & \textbf{constrained}                                                                                                                                                                                                                                                                                                                                          & \textbf{unconstrained}                                                                                                                                                                                                                                                                                    \\ \midrule
\midrule
Slaughter fought for law and order with his six-shooter, a shotgun, and a repeating Henry rifle. He arrested desperados like the Jack Taylor Gang and brought them to justice.He also became a prominent poker player, often playing all night long. He was reportedly good at bluffing. He often played with the cattle baron John Chisum.                                         & \{('Slaughter\_(band)', 'armament', 'Shotgun'), ('Slaughter\_(band)', 'armament', 'Rifle'), ('Slaughter\_(band)', 'armament', 'Henry\_rifle')\}                                                                                                                                                                                                      & \{('Slaughter\_(poker)', 'armament', 'Henry\_rifle'), ('Slaughter\_(poker)', 'armament', 'Shotgun'), ('Slaughter\_(poker)', 'armament', 'Six-shooter'), ('Slaughter\_(poker)', 'unmarried partner', 'John\_Chisum')\}                                                                            \\
\midrule
Carol Dollard, who once worked in product development for PepsiCo, told Gladwell: "I've seen many times where the sip test will give you one result and the home-use test will give you the exact opposite.". For example, although many consumers react positively to the sweeter taste of Pepsi in small volumes, it may become unattractively sweet when drunk in quantity. & \{('Carol\_Douglas', 'instance of', 'Human'), ('PepsiCo', 'industry', 'Food\_industry'), ('PepsiCo', 'product or material produced', 'Pepsi'), ('Carol\_Douglas', 'employer', 'PepsiCo')\}                                                                                                                                                           & \{('Carol\_Dollard', 'employer', 'PepsiCo'), ('Carol\_Dollard', 'described by source', 'Gladwell\_Encyclopedic\_Dictionary'), ('PepsiCo', 'product or material produced', 'Pepsi')\}                                                                                                             \\
\midrule
In June 1998, the founders became unhappy with the direction in which 3Com was taking the company, and left to found Handspring.                                                                                                                                                                                                                                                    & \{('3Com', 'followed by', 'Handspring\_(company)')\}                                                                                                                                                                                                                                                                                                 & \{('3Com', 'replaced by', 'Handspring\_(company)'), ('3Com', 'significant event', 'June\_1998')\}                                                                                                                                                                                                \\
\midrule
The merger also brought the game download site FilePlanet into the IGN group; as of 2011 both FilePlanet and the GameSpy website still operate as video game-related web sites.                                                                                                                                                                                                     & \{('Mergers\_and\_acquisitions', 'facet of', 'Software\_as\_a\_service'), ('Mergers\_and\_acquisitions', 'facet of', 'Software'), ('Video\_game\_industry', 'product or material produced', 'Video\_game'), ('Video\_game\_industry', 'instance of', 'Industry\_(economics)'), ('Mergers\_and\_acquisitions', 'facet of', 'Video\_game\_industry')\} & \{('Merger\_of\_FilePlanet\_and\_GameSpy', 'business division', 'FilePlanet'), ('Merger\_of\_FilePlanet\_and\_GameSpy', 'business division', 'GameSpy')\}                                                                                                                                        \\
\midrule
Keith Taylor and Mike Scarrott called it the "Boomerang Nebula" in 1980 after observing it with the Anglo-Australian telescope at the Siding Spring Observatory. Unable to view it with great clarity, the astronomers saw merely a slight asymmetry in the nebula's lobes suggesting a curved shape like a boomerang.                                                              & \{('Boomerang\_Nebula', 'astronomical filter', 'Visual\_perception'), ('Boomerang\_Nebula', 'parent astronomical body', 'Sun'), ('Visual\_perception', 'subclass of', 'Perception')\}                                                                                                                                                                & \{('Boomerang\_Nebula', 'site of astronomical discovery', 'Siding\_Spring\_Observatory'), ('Boomerang\_Nebula', 'named after', 'Boomerang'), ('Boomerang\_Nebula', 'discoverer or inventor', 'Keith\_Taylor\_(astronomer)'), ('Boomerang\_Nebula', 'discoverer or inventor', 'Mike\_Scarrott')\} \\ \bottomrule
\end{tabular}
}
\caption{Examples of generation with SynthIE in constrained and unconstrained mode}
\label{tab:synthie-constrained-unconstrained}
\end{table*}
\begin{table*}[]
\centering
\resizebox{\textwidth}{!}{
\setlength{\tabcolsep}{5pt}
\begin{tabular}{@{}p{0.45\textwidth}|p{0.45\textwidth}|p{0.45\textwidth}@{}}
\toprule
\textbf{Text}                                                                                                              & \textbf{\relik{}}                                                                                                                                                                               & \textbf{SynthIE (unconstrained)}                                                                                                                                                                                                                                                                \\
\midrule
\midrule
The Verwall Alps are not a mountain range in Austria's Vorarlberg region, which borders the Samnaun Alps.         & {[}{[}'Verwall\_Alps', 'country', 'Austria'{]}, {[}'Vorarlberg', 'country', 'Austria'{]}, {[}'Vorarlberg', 'location', 'Austria'{]}, {[}'Samnaun\_Alps', 'country', 'Austria'{]}{]} & {[}{[}'Verwall\_Alps', 'different from', 'Vorarlberg'{]}, {[}'Vorarlberg', 'shares border with', 'Samnaun\_Alps'{]}{]}                                                                                                                                                                \\
\midrule
Windows Nashville was not a codename for a cancelled release of Microsoft Windows.                                & {[}{[}'Windows\_Nashville', 'edition or translation of', 'Microsoft\_Windows'{]}{]}                                                                                                 & None                                                                                                                                                                                                                                                                                  \\
\midrule
"The Land of Mist" is not a fantasy short story published in the Strand Magazine. It is not in the public domain. & {[}{[}'The\_Land\_of\_Mist', 'published in', 'The\_Strand\_Magazine'{]}{]}                                                                                                          & {[}{[}'The\_Land\_of\_Mist', 'different from' 'Fantasy\_short\_story'{]}, {[}'The\_Land\_of\_Mist', 'published in', 'Strand\_Magazine'{]}, {[}'The\_Land\_of\_Mist', 'different from', 'The\_Land\_of\_Mist'{]}, {[}'The\_Land\_of\_Mist', 'copyright status', 'Public\_domain'{]}{]} \\
\midrule
Münchner Illustrierte is not a German magazine.                                                                   & {[}{[}'Münchner\_Illustrierte', 'instance of', 'Magazine'{]}{]}                                                                                                                     & None                                                                                                                                                                                                                                                                                  \\
\midrule
"Groovin' Blue" is not an album by Curtis Amy, released on Pacific Jazz Records.                                  & {[}{[}'Groovin\textbackslash{}'\_Blue', 'performer', 'Curtis\_Amy'{]}, {[}'Curtis\_Amy', 'record label', 'Pacific \_Jazz\_Records'{]}{]}                                            & {[}{[}'Groovin\textbackslash{}'\_Blue, 'different from', 'Groovin'\_Blue\_(Curtis\_Amy\_album)'{]}, {[}'Groovin\textbackslash{}'\_Blue', 'record label', 'Pacific\_Jazz\_Records'{]}{]}       \\
\bottomrule
\end{tabular}
}
\caption{Examples of outputs from \relik{} and SynthIE on negated \synthiedata{} data samples}
\label{tab:negation}
\end{table*}

\subsection{Constrained \vs unconstrained generation}
\label{appendix:constrained_unconstrained_analysis}

As mentioned in \Secref{sec:method}, we suspect that errors that stem from the constrained and unconstrained decoding are complementary (see \Figref{fig:diagram}). To support this claim, we perform an analysis of the errors that occur during constrained and unconstrained decoding. We use the same error classification and Wikipedia data as in \Secref{sec:error_analysis}. We also add another type of error that only happens during the unconstrained decoding -- \textbf{Entity/relation not in the KB} -- which covers the case for which generated entities or relations are not present in the knowledge base. The results are shown in \Tabref{tab:synthie_constrained_unconstrained_analysis}.

In \Tabref{tab:synthie-constrained-unconstrained}, we show examples of SynthIE outputs in both constrained and unconstrained manner, on real Wikipedia text. Overall, SynthIE does open information extraction well (i.e. without KB constraints), but constrained decoding only works when there are not many deviations between facts in the text and the KB.

\subsection{Analysis with negation}

We suspect that, since \relik{} was trained to match retrieved entities and relations with spans of text identified as relevant, it is more likely to find a relation between two entities in the text that are not connected. To confirm this, we edit samples from the test split of \synthiedata, created by \citet{synthie} for the SynthIE model, by replacing ``is'' in text with ``is not''. For example: ``Groovin' Blue is not an album by Curtis Amy, released on Pacific Jazz Records'' would be used instead of ``Groovin' Blue is an album by Curtis Amy, released on Pacific Jazz Records''. We test both SynthIE (without constrained decoding) and \relik\ on this modified data. Neither of the two models performs well on this task, but there are indicators that SynthIE is able to somewhat model the lack of relation between two entities. In the case of \relik, this happens rarely. For examples, see \Tabref{tab:negation}

\subsection{GPT-4o pipeline}
\label{appendix:gpt4o}
\begin{table*}[]
\centering
\resizebox{\textwidth}{!}{
\setlength{\tabcolsep}{5pt}
\begin{tabular}{@{}p{0.45\textwidth}|p{0.45\textwidth}|p{0.45\textwidth}@{}}
\toprule
\textbf{Text}                                                                                                                                                                                                                                                                                                                                                                                                                                                                                                                                                                                                              & \textbf{GPT-4o }                                                                                                                                                                                                                                                                  & \textbf{GPT-4o + filter by entities}                                                                               \\ \midrule
\midrule
A film adaptation of Asada's work, directed by Yōjirō Takita, is known as When the Last Sword Is Drawn. A ten-hour 2002 television jidaigeki based on the same novel starred Ken Watanabe. A Japanese Manga Series, Rurouni Kenshin, by Nobuhiro Watsuki, notably sets place in the war, and the aftermath. Western interpretations include the 2003 American film The Last Samurai directed by Edward Zwick, which combines into a single narrative historical situations belonging both to the Boshin War, the 1877 Satsuma Rebellion, and other similar uprisings of ex-samurai during the early Meiji period. & ('Rurouni\_Kenshin', 'creator', 'Nobuhiro\_Watsuki'), ('The\_Last\_Samurai', 'director', 'Edward\_Zwick')                                                                                                                                                                & ('Rurouni\_Kenshin', 'creator', 'Nobuhiro\_Watsuki'), ('The\_Last\_Samurai', 'director', 'Edward\_Zwick') \\
\midrule
Carol Dollard, who once worked in product development for PepsiCo, told Gladwell: "I've seen many times where the sip test will give you one result and the home-use test will give you the exact opposite.". For example, although many consumers react positively to the sweeter taste of Pepsi in small volumes, it may become unattractively sweet when drunk in quantity.                                                                                                                                                                                                                               & ('PepsiCo', 'product or material produced', 'Pepsi')                                                                                                                                                                                                                     & ('PepsiCo', 'product or material produced', 'Pepsi')                                                      \\
\midrule
At the time he ran a chain of health food shops in the United States, with which she became involved. The couple now live in London and have two daughters.                                                                                                                                                                                                                                                                                                                                                                                                                                                       & ('John\_Henry\_Richardson', 'residence', 'London')                                                                                                                                                                                                                       & None                                                                                                      \\
\midrule
However, the U.S. Bank Tower does surpass both the Salesforce Tower and the Wilshire Grand Center in roof height, making it the only building in California whose roof height exceeds 1,000 feet (300 m). Because local building codes required all high-rise buildings to have a helipad, it was known as the tallest building in the world with a rooftop heliport from its completion in 1989 to 2010 when the China World Trade Center Tower III opened.                                                                                                                                                      & ('Wilshire\_Grand\_Center', 'roof height comparison', 'Salesforce\_Tower'), ('U.S.\_Bank\_Tower', 'located in', 'California'), ('U.S.\_Bank\_Tower', 'has part', 'Heliport'), ('U.S.\_Bank\_Tower', 'heliport timeframe end', 'China\_World\_Trade\_Center\_Tower\_III') & ('Wilshire\_Grand\_Center', 'roof height comparison', 'Salesforce\_Tower')                                \\
\midrule
Thorpe immediately is enchanted by Doña María and gallantly returns her plundered jewels. Her detestation of him softens as she too begins to fall in love.                                                                                                                                                                                                                                                                                                                                                                                                                                                       & None                                                                                                                                                                                                                                                                     & None                                                                                                      \\ \bottomrule
\end{tabular}
}
\caption{Examples generated by GPT-4o pipeline. Second column presents raw outputs after being prompted with our pipeline. Third column presents results where triplets containing entities which are not in the retrieved entities are removed.}
\label{tab:gpt-4o-real}
\end{table*}

In \Secref{sec:error_analysis}, we show some of the disadvantages of the current approaches with the smaller LMs. However, LLMs are more powerful in terms of their external knowledge, which can be a useful thing when extracting information facts. The pitfall with LLMs for this task is the KB. As they do not possess information about what is present in our KB, they are struggling to output the triplets in the correct format, or under correct constraints.

Under the assumption that one has unlimited resources for this task, we tried using GPT-4o with a form of retrieval-augmented generation (RAG). In this way, the LLM has the information about our KB. Here we present some of the key improvements to the standard prompt that resulted in better outputs (manually evaluated):

\begin{itemize}
    \item \textbf{Entity retrieval}: We noticed that it is important for entity retrieval to be high-recall. This means that we did not care if many entities were not relevant, as long as all the relevant ones were included. GPT-4o seems to be able to filter the non-relevant entities, but cannot come up with the new ones. In our case, we used a mix of entities retrieved by \relik, SynthIE, and GenIE (both in an unconstrained setting). We did not include relation retrieval as we find this to be a harder task than entity retrieval, which requires the model to almost be able to do cIE on its own. Theoretically, with LLMs that have longer context sizes, in our case, it is possible to send the whole list of relations. We did not test this but expect that this would improve the performance.
    \item \textbf{Sketch of triplet generation}: We noticed that GPT-4o produces better outputs when a sketch of a triplet generation by some other model is provided. Anecdotally, the outputs were better even when the sketches were bad. For the sketch, we used the output of the SynthIE model
    \item \textbf{Encourage reasoning}: LLM was performing vastly better when it was encouraged to explain the reasoning behind the choice of the triplets
\end{itemize}

We did not perform a formal evaluation of this method as it was not the focus of our study. All our findings from this section are based on manual inspection of the results. One thing we draw attention to is that LLMs have likely been exposed to the data we used for our manual inspection during their pretraining. Second thing to be careful about are rare relations. As they do not appear often, it is likely that an LLM would prioritize more common relations when generating the output. Regardless of that, we showcase our attempt as a starting point for the other researchers. For examples of generated outputs with GPT-4o on the real Wikipedia data, see \Tabref{tab:gpt-4o-real}.

\end{document}